\documentclass{llncs}
\usepackage{llncsdoc}
\usepackage{graphicx}
\usepackage{algorithmic}
\usepackage{algorithm}
\usepackage{color}

\graphicspath{ {images/} }

\begin{document}

\title{Substation Signal Matching with a \\ Bagged Token Classifier}

\author{Qin Wang\inst{1} \and Sandro Sch\"onborn\inst{2} \and Yvonne-Anne Pignolet\inst{2} \and \\Theo Widmer\inst{3} \and Carsten Franke\inst{2}}

\institute{ETH Z\"urich, \email{qwang@student.ethz.ch}
\and ABB Corporate Research, \email{\{firstname.lastname\}@ch.abb.com}
\and ABB Power Grid - Grid Automation, \email{theo.widmer@ch.abb.com}}
\maketitle


%
\begin{abstract}
Currently, engineers at substation service providers match customer data with the
corresponding internally used signal names manually. 
This paper proposes a machine learning method to automate this process based on substation signal mapping data from a repository of executed projects. To this end, a bagged token classifier is proposed, letting words (tokens) in the customer signal name vote for provider signal names. 
In our evaluation, the proposed method exhibits better performance in terms of both accuracy and efficiency over standard classifiers.
\end{abstract}

\section{Introduction} 
\label{sec:introduction}
Matching utility customer-specified signal names for protection, control and monitoring functions with signal names used by a system provider is a common task in substation automation engineering. To ensure consistency, the system providers maintain an internal library that contains the names to be used for function signals for all projects. This helps the system provider to standardize and streamline its processes and ensures that signal names of important substation automation functions are used in the same manner. On the other hand, the naming schemes used by customers usually differ, both among different customers and compared to provider libraries. 
Consequently, when starting to work on a new substation automation project, an engineer at the system provider must assign the correct library signal names to customer signal names, a cumbersome, error-prone, and time-consuming process. The matching quality is extremely important to ensure the substation automation systems can work correctly and fits in the customer's environment once deployed and the customer's tools can interoperate with it seamlessly. Hence, in current practice this task is typically carried out by experienced engineers.

The objective of our paper is to find a way to automate this process and thus to improve the engineers' efficiency. More precisely, we  present how we devised and evaluated a machine learning-based system that suggest matching library signal names for customer-specified signal names.
%
%
The system extracts its internal knowledge from past projects that were carried out with a manual signal name assignment. In other words, a repository of past projects is used to build training and testing data sets for our system. 
Signal name matching is difficult as customer signal names can be arbitrary and typically contain abbreviations, ambiguity and misspellings. Different naming schemes for lines, e.g., \texttt{L1, L2, L3} or \texttt{R, Y, B} are both used. In contrast, the provider library consists of a restricted set of known unique and clean signal names. In past project data, the engineers that carried out the matching sometimes made mistakes or ignored the library signal names, in other words the learning data is noisy. 

Since the signal names in the provider library are fixed, this matching problem can be modeled as a text classification task predicting library signal names for customer signal names. Text classification has been well explored by the machine learning community~\cite{aggarwal2012survey}. Name matching can also be approached by string comparison with flexible and even adaptive string distance metrics~\cite{bilenko2003adaptive}. Such methods are targeted more towards very similar strings with occasional differences such as misspelled words, missing parts etc. In our case, library and customer names are not necessarily related in the choice of words they use. Schema matching also deals with identifying the same entities in two sets of names~\cite{rahm2001survey}. It focuses on matching complete schema consisting of many names organized in a structure, usually for record linkage in databases. In our case, we do not expect the various substation setups to adhere to similar structures and thus consider each name individually. By choosing simple text classification, solely based on customer signal name, we neither require textual similarity between customer and library names nor a common structure among signal names.

We propose to use an efficient and accurate token dictionary as a name classifier. In the token dictionary, each word of the customer signal name can vote for possible library names. It is similar to a Na\"ive Bayes classifier but aggregates and normalizes votes differently. We also explore and adapt a range of existing text classification methods and compare their classification performance as well as their computational efficiency. For this evaluation, we consider standard text classification methods, such as Na\"ive Bayes, Random Forest and Support Vector Machine and additionally construct a sequence-aware recurrent neural network.

The paper is organized as follows. In Section~\ref{sec:problem}, we introduce the problem formally with details, followed by a high-level system description in Section~\ref{sec:overview}. In Section~\ref{sec:methods}, different classification methods along with our proposed methods are presented. We describe our evaluation in Section~\ref{sec:evaluation}. Conclusions are given in Section~\ref{sec:conclusion}.


\section{Problem description}
\label{sec:problem}

\paragraph{Classification Problem}
The selection of provider signal names from customer signal names is modeled as a classification problem, as the signal names are always chosen from the limited number of possible provider signal names. To this end, each possible provider signal forms its own class. Therefore, the problem is an instance of multi-class classification. In our case, we use 3745 possible classes in the provider signal library. 
The formal task is to \emph{predict} the correct signal class $c$ for a given customer input name $\tilde x$, encoded as a string. A model for the prediction is learned from past project data. To support the engineer with multiple possibilities to choose from, the system is to suggest $k$ candidates of matching provider names $c_1, c_2, \ldots, c_k$, sorted by their relevance.

\noindent\paragraph{Dataset}
~\label{sec:dataset}
The data set that is used for this work consists of totally 8969 unique pairs of customer signal names with corresponding provider signal names from 170 past projects. Projects have a varying degree of similarity. The overlap across customer signal names is rather low between different projects, indicating the need to standardize names using library signal names.



%
\section{System Overview}
\label{sec:overview}

We setup and evaluate a machine learning pipeline with different classification algorithms to identify library signals from input customer signal names. The pipeline consists of methods for data pre-processing, classification and post-processing. Pre-processing prepares the raw input so it can be processed by the classification algorithms applied afterwards. In particular, they require tokenized text or numerical vectors as input. We select and compare multiple algorithms with respect to different criteria and performance. In post-processing, we ensure certain hard constraints, such as respecting known antonyms in the final signal name. 

\begin{figure}
\center
	\vspace{-.5 cm}
	\includegraphics[width=0.95\textwidth]{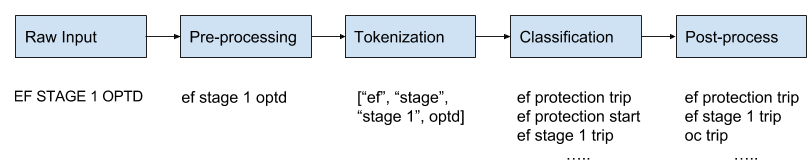}	
	\vspace{-.5 cm}
	\caption[ ]{Proposed pipeline to identify library signals from input customer signal names}
	\label{fig:pipeline}
\end{figure}
\vspace{-1cm}

\subsection{Pre-processing}

\label{sec:removeUnmatchable}

Pre-processing consists of two steps data cleaning, normalization and tokenization. Data cleaning is mainly relevant to build a good training set while normalization and tokenization are always applied for new customer input signal names, also in test scenarios.


\paragraph{Normalization} Signal names are normalized to lower case. This avoids mismatches due to different capitalization methods.

\paragraph{Cleaning} 
Signal name pairs where the provider name  does not occur in the library are  removed. These cases exist because either special naming schemes were required by the customer, or the project was implemented before the library was created. Furhtermore, we remove all examples that have identical customer and internal signal name. Theses examples can easily be recognized and predicted, and are thus not our main interest and would not help to discriminate between different methods in the evaluation. After all the normalization and cleaning steps, we have a dataset appropriate for learning and testing.


\paragraph{Tokenization}
\label{token}
Our methods rely on tokens extracted from signal names for classification. A token is typically a single word in a multi-word signal name. We split between words using a set of separator characters. The only exception are a number following a noun. In this case, we tokenize \texttt{Noun N} into \texttt{Noun} and \texttt{Noun N} to capture the context of the number. Empty tokens are discarded.

Customer signal names are then represented as a vector of token counts, similar to the bag-of-words model~\cite{hotho2005brief}. A dictionary of known tokens is extracted from the training set. For some methods, we use 3-grams representation where we add all 3-grams to the set of tokens.

\begin{table}
\vspace{-.5cm}
	\begin{center}
		\renewcommand{\arraystretch}{1.4}
		\setlength\tabcolsep{3pt}
		\begin{tabular}{llllll}
			\hline\noalign{\smallskip}
			Original Customer Signal & Processed Customer Signal\\
			\noalign{\smallskip}
			\hline
			\noalign{\smallskip}
			Dist. Zone 2 Trip & ["dist", "zone", "zone 2", "trip"] \\
			CR\&WEI Dist. Rev Log. Blocked & ["cr", "wei", "dist", "rev", "log", "blocked"]  \\
			Block (B Inhibit) automatic control & ["block", "b", "inhibit", "automatic", "control"]\\
			\hline
		\end{tabular}
	\end{center}
\caption{Pre-processing Examples.}
	\end{table}

\vspace{-1.4cm}

\subsection{Classification}
Classification algorithms take the tokenized signal as an input and identify the best matching library signal classes. 
We evaluated a few basic choices for text classification such as Na\"ive Bayes, Random Forest, and linear Support Vector Machine (SVM). Furthermore, we present a bespoke Neural Network approach and devise a dictionary approach, called token dictionary, which turns out to be optimal for this problem in terms of performance and hardware demands. A very basic lookup table serves as a baseline for all methods. The methods are also required to make multiple predictions such that the engineer can choose from multiple matches. More details are presented in the next section.

\subsection{Post-processing}
\label{sec:postprocessing}
For substation signal mapping, there are antonym token pairs that should not appear simultaneously on customer and internal signal names. For example, if ``underfreq'' is a token in the customer signal name, then the prediction should never contain ``overfreq''. In addition, there are also keywords which must appear on both customer and internal sides. For example, if ``interlocked'' is a token in the customer signal name, then the prediction should also contain this key token. In order to make our predictions more accurate, a post-processing pipeline is implemented to manually penalize predictions which contain an antonymous token and reward those with the same keywords as customer signal. This processing step basically reorders our list of predictions to make sure that better predictions are shown on top.
Pseudocode for the algorithms re-ranking the predictions can be found in 
Algorithm~\ref{alg:post} .

\begin{algorithm}[h]
	\caption{Post-processing Procedure}
	\begin{algorithmic}[1]
	  \REQUIRE customerSignal, predictionList, antonymDict, keywordSet 
		\COMMENT{\\ *** penalize occurence of forbidden words **}
		\STATE matchedTokens = antonymDict.keys.intersection(customerSignal.tokens)
		\STATE forbiddenWords = flatten([antonymDict[key] for key in matchedTokens])
		\STATE \textbf{for} {prediction in predictionList} \textbf{do}
		\STATE ~~~~\textbf {if} {forbiddenWords.intersection(prediction.tokens) is not empty} \textbf{then}
		\STATE ~~~~ ~~~~Move prediction to the end of predictionList
		\COMMENT{\\ *** reward occurence of keywords **}
		\STATE matchedTokens = keywordSet.intersection(customerSignal.tokens)
		\STATE \textbf{for} {prediction in predictionList.reverse} \textbf{do}
		\STATE ~~~~\textbf{if} {matchedTokens.intersection(prediction.tokens) is not empty} \textbf{then}
		\STATE ~~~~ ~~~~Move prediction to the front of predictionList
	\end{algorithmic}
	\label{alg:post}
\end{algorithm}

%
\section{Classification Methods}
\label{sec:methods}
In this section we present the core classification methods we investigate for our system. We start by a description of well-known standard methods, followed by the approaches we devised specifically for this problem, an LSTM neural network approach and a token dictionary approach.

\subsection{Standard Classification Methods}

\paragraph{Lookup Table}
We consider a simple lookup table as our baseline. For each customer signal that appeared in the training set, we maintain a list of corresponding library signals. The list is sorted by appearance frequency given the customer signal name in the training set. Given a test customer signal, the table returns a sorted list of up to $k$ library signals. 

\paragraph{Na\"ive Bayes}
The Na\"ive Bayes method assumes conditional independence among 
multinomial token occurrence probabilities for each class. Despite the simplifying assumption, it often works CF: surprisingly is not very scientific: surprisingly well for real-world text classification~\cite{mccallum1998comparison}. A significant advantage of Na\"ive Bayes classifiers over other sophisticated methods is that they require a small amount of training data and can be trained very efficiently. The resulting model typically provides fast classification with a moderate memory footprint. We use the Na\"ive Bayes implementation of scikit-learn~\cite{pedregosa2011scikit} with our bag-of-words token count vectors as features. Since Na\"ive Bayes classifiers calculate posterior probabilities for all classes, we are able to select the best k predictions as those with highest posterior probability.

\paragraph{Random Forest}
Random forests are ensemble learning methods that counteract single decision tree's shortcomings by taking many trees into account~\cite{liaw2002classification}. Random forests are generic classifiers which can be applied to many problems with high success rates~\cite{caruana2006empirical}. Contrary to Na\"ive Bayes classifiers, they can also represent (non-sequential) dependencies among tokens. Random Forests are efficient to train and classify but can require a lot of memory to store all the trees.

We again use the scikit-learn default implementation with token vectors as input. A few very deep trees performed better than many shallow trees. We restricted the number of trees to 10 and left the depth unconstrained, which resulted in depths up to 604. Random forests can report classification certainties which serve us to select best $k$ matches.

However, random forests are not optimal for our problem. Due to the large vocabulary size but short signal length, the features are sparse. Bagging and suboptimal selection of splits can waste random forest's effort on zero-areas.

\paragraph{Support Vector Machine}
Most text categorization problems are linearly separable~\cite{joachims1998text} and Random forests are not optimal for very high dimensional sparse feature vectors. Thus linear Support Vector Machine (SVM) classifiers that maximize margins are widely used for text classification. SVM training is typically memory-inefficient for large datasets and sparse features. We thus resort to the stochastic linear Hinge loss SVM described in \cite{bottou2010large} to reduce computation time and memory footprint, using the scikit-learn implementation. Memory consumption of this SVM model is typically moderate as well as the time required for training and classification. We split part of the training set for probability calibration, which enables us to assign probability to each class and suggest $k$ candidate signal names.

\subsection{Recurrent Neural Network}

\begin{figure}
	\includegraphics[width=0.95\textwidth]{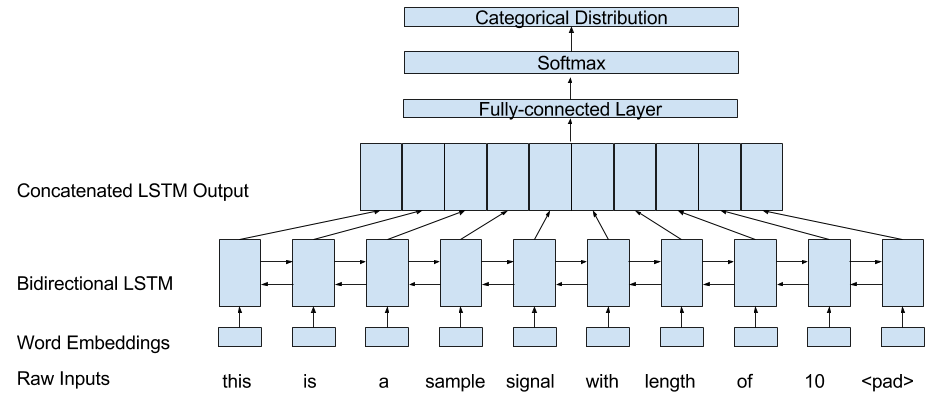}	
\vspace{-.2cm}
	\caption[ ]{Our BLSTM classification network.}
	\label{fig:lstm}
\end{figure}

All methods based on bag-of-words tokens ignore the order of words in a signal name. To capture the sequence of tokens in a signal name, we implement a recurrent neural network of the LSTM-type to classify a sequence of ``GloVe'' embedding vectors of individual tokens. Such a setup has been successfully applied to large text classification problems but requires a lot of resources~\cite{tang2015document}.
 
\paragraph{Vectorization and Embedding}
Unlike the above models that use bag-of-word counts as input features, we use pre-trained GloVe~\cite{pennington2014glove} word embeddings in our neural model. Embeddings are pre-trained dictionaries that map tokens to vectors in a linear space, where words with similar meanings lie close to each other. In our case, the 50-dimension GloVe embeddings are used. To capture the sequence, a customer signal is always first cropped or padded to ten tokens. All tokens are then mapped to vectors using the embedding lookup table. Note that the embedding lookup table is initialized by pre-trained GloVe embeddings but is still trainable during the optimization process.


\paragraph{Classification network}
Bidirectional Long Short-Term Memory (BLSTM) networks have been shown to outperform unidirectional LSTMs, standard Recurrent Neural networks, and Multilayer Perceptrons on text tasks due to their stronger ability to capture contextual information~\cite{graves2005framewise}. In our model, one BLSTM layer processes the sequence of 10 token vectors as input and generates 10 output vectors.  We choose a fixed length of 10 because most customer signals contain fewer than 10 tokens. In the LSTM cell, we use the standard sigmoid activation and a hidden size of 64 units. All output vectors are concatenated and fed to a fully connected classification layer with a softmax output providing the predicted class probabilities. The fully connected layer learns a linear transform of the LSTM output to class scores. Softmax interprets class scores as unnormalized log probabilities for each class and normalizes them appropriately~\cite{bishop2006pattern}.


\paragraph{Optimization}
We train the LSTM classifier and all network components by minimizing the cross-entropy loss of softmax logits using Adam Optimizer~\cite{kingma2014adam}. A fixed learning-rate of 1e-3 is used. Network and optimization algorithm are implemented using TensorFlow~\cite{abadi2016tensorflow}. 

\subsection{Token Dictionary}

A lookup of the complete signal name, as with the lookup table, is too specific and does not generalize well. But typical customer names still contain specific words which indicate the appropriate library name, almost like keywords. We thus introduce a token dictionary which looks up each token individually and lets it vote for library names it appeared with in the past. Voting allows for ambiguity where the same keyword appears in many classes. Each token votes for all possible hypotheses that could have generated it.

The test signal $\tilde x$, to be classified, is treated as a set of its $N$ tokens $\tilde x = [t_1, t_2, \ldots, t_N]$. Each token $t_i$ votes for all classes  according to the frequency of co-occurrence. The vote of a token $t_i$ for class $c$ is its number of occurrences in examples for said class $n(t_i, c)$. The weight of a vote for class $c$ given token $t_i$ is computed from $n(t_i, c)$, the frequency the token appeared in samples of class~$c$.

\vspace{-.2cm}
\begin{equation}
v(c \mid t_i) = \frac{n(t_i, c)}{\sum_{c'} n(t_i, c')} = \frac{n(t_i, c)}{n(t_i)}.
\end{equation}

By normalization, the total vote of a single token is split among all possible classes. Common tokens will only contribute weak votes compared to more discriminative tokens. This effect is similar to the one achieved by inverse document frequency normalization. All token votes are summed to form the total vote for for the complete customer name.

\vspace{-.2cm}
\begin{equation}
v(c \mid \tilde{x}) = \sum_{i=1}^{N} v(c \mid t_i).
\end{equation}	

The normalized votes form a probability distribution over all possible $K$~classes.

\vspace{-.2cm}
\begin{equation}
\label{eq:singletokenposterior}
P(c \mid \tilde{x}) = \frac{v(c \mid \tilde{x})}{\sum_{c'=1}^K v(c' \mid 
	\tilde{x})}.
\end{equation}

Formally, such a classifier is a bagged collection of discriminative, weak, single-token classifiers. The token dictionary classifier works as a bag-of-words model. The resulting vote aggregation adds individual contributions and is thus different from multiplying token likelihoods $P(t_i \mid c)$ in the Na\"ive Bayes classifier. Also, consider the different normalization of token likelihoods and single-token posterior~(\ref{eq:singletokenposterior}). Aggregation of additive votes typically leads to broader prediction distributions than in the Naive Bayes case. Also, adding votes ensures that a single token can "activate" a library name while all other typical words are absent. To ensure such behavior, the Na\"ive Bayes method needs explicit prior initialization, e.g. with Laplace smoothing.

By choosing the top $k$ classes with maximal $P(c \mid \tilde{x})$, the algorithm can be easily extended to multiple prediction cases. Our token dictionary implementation uses all 3-grams as input, including single tokens and 2-grams. It is based on standard python hashtables and thus very light in memory and extremely fast at learning and predicting.


%
\section{Evaluation}
\label{sec:evaluation}
%
All evaluation experiments are run on a workstation with a 4.4.0 Linux kernel, an Intel(R) Core(TM) i5-4570 CPU @ 3.20GHz processor (four cores), and 8GB of RAM memory. Scikit-learn and Tensorflow are the main libraries we used. For the comparison, results of Lookup Table, Naive Bayes, Random Forest, SVM, LSTM, and our proposed token dictionary are shown in the next chapter.
The models are evaluated on the dataset described in Section~\ref{sec:dataset}, including all tokens appearing in the library. 
We randomly select 34 (20\%) of the projects as the test set and use the rest for training.


We evaluate the different methods using the following metrics:

\textit{Classification Performance}
We report accuracy, weighted recall and weighted F1 of the models. In addition, \texttt{Top 10 accuracy} is provided. In this metric, the prediction list for a single query is considered as a match if the true label appears in the list. This loosens the requirement and better reflects how efficient it can be when using the software in a production scenario where engineers look through the list of suggestions to determine the best choice. 

In order to know the data requirement for our models, we compare the accuracy of methods when different amounts of data are used in training. This will give us an estimate of how much data is necessary for different methods.

\textit{Run Time Evaluation}
A runtime comparison evaluates the training time on the whole training set as well as the prediction time per query.

\textit{Memory Usage Evaluation}
In addition, we compare the memory consumption of methods under scrutiny. In addition to peak memory usage during training, the model size is provided as an indication of memory usage for predictions. 


%
%
\subsection{Classification performance}
All testing models outperform our baseline look-up table by a considerable margin, showing that the engineers' efficiency can be improved by using machine learning methods. For single-prediction results, random forest outperforms all the other models in term of accuracy, F1, and Recall. In terms of top 10 accuracy, the proposed token dictionary and Na\"ive bayes outperform other models by at least 5\% and achieve 91\% accuracy. This means these two methods offer suggestions of higher quality. LSTM performs worse than other classifiers despite its large model size, indicating that temporal token dependencies are not crucial in our problem.


\begin{table}[h]
\vspace{-.4cm}
	\begin{center}
		\renewcommand{\arraystretch}{1.4}
		\setlength\tabcolsep{3pt}
		\begin{tabular}{llllll}
			\hline\noalign{\smallskip}
			Method & Accuracy & Top 10 Accuracy & F1
			& Recall\\
			\noalign{\smallskip}
			\hline
			\noalign{\smallskip}
			Lookup Table & 0.66 & 0.74 & 0.67 & 0.66 \\
			Naive Bayes & 0.70 & \textbf{0.91} & 0.69 & 0.70 \\
			Linear SVM & 0.69 & 0.90 & 0.69 & 0.69 \\
			LSTM & 0.68 & 0.85 & 0.68 & 0.68 \\
			Random Forest & \textbf{0.78} & 0.88 &  \textbf{0.79} &  \textbf{0.78} \\
			Token Dict & 0.73 &  \textbf{0.91} & 0.73 & 0.73 \\
			\hline
		\end{tabular}
	\end{center}
	\caption{Evaluation results based on full training set.}
\end{table}

\begin{figure}
	\vspace{1 cm}
	\includegraphics[width=0.95\textwidth]{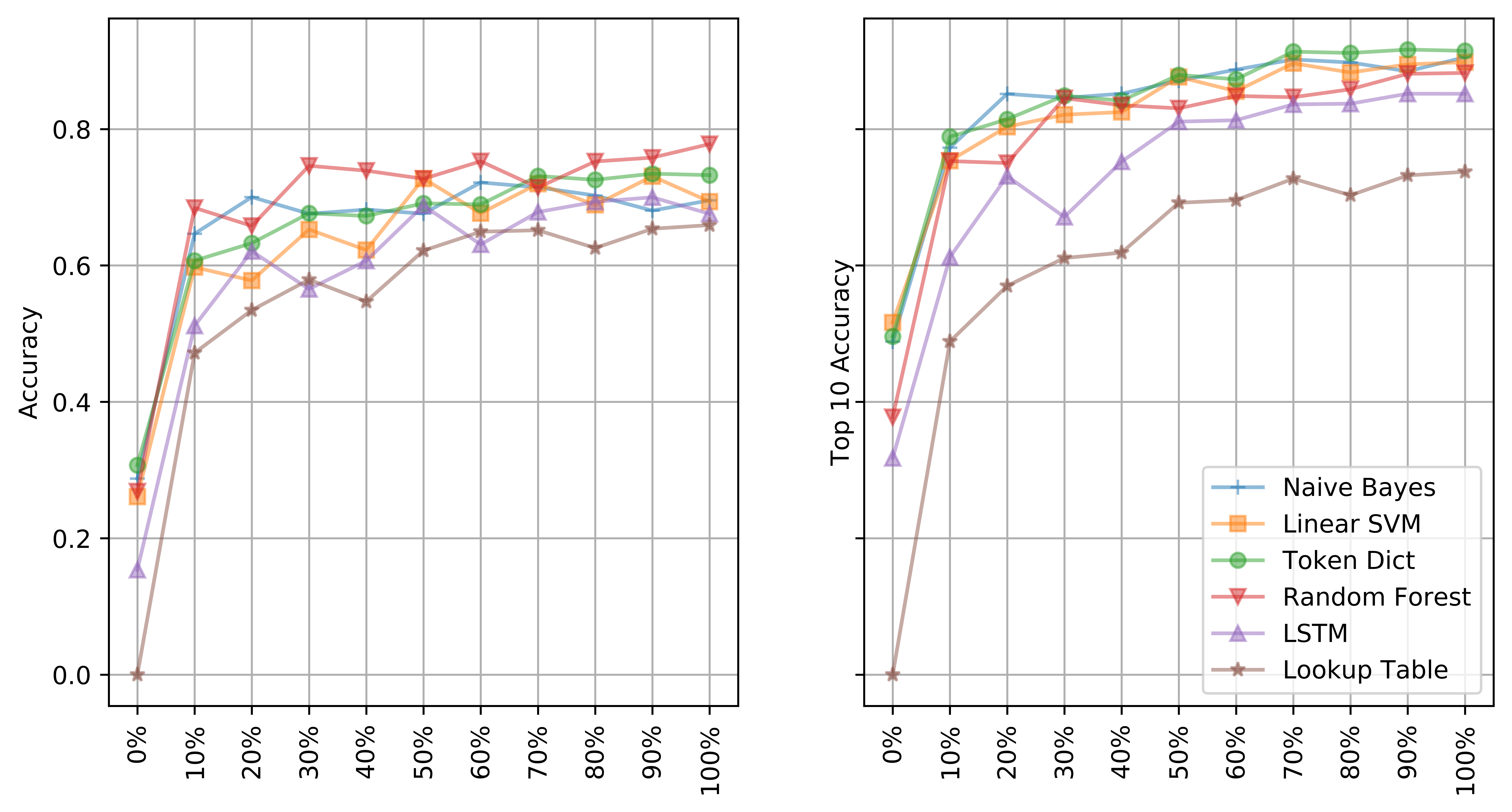}	
	\caption[ ]{Prediction accuracy against amount of training data.}
	\label{fig:accuracy}
\end{figure}

In addition to standard evaluation methods, the influence of the amount of training data on accuracy is measured. As shown in Figure~\ref{fig:accuracy}, in terms of accuracy, most methods continuously improve when more data are fed. However, this improvement is not significant: less than 5\% difference is achieved when when using 100\% instead of 50\% training data.

\subsection{Run Time}
Table~\ref{tab:runtime} presents the runtime measurements of the training and testing phase, using the whole training set for all models.
In terms of training time, LSTM is significantly slower than the others because of its large number of parameters to train. SVM can be trained within ten minutes. The other three models, Naive Bayes, Random Forest, and Token Dictionary execute the training within less than one minute.
In terms of prediction time, all models except SVM respond to each query within 0.1s on average, indicating that these models can be directly used by engineers on a standard workstation without inflicting a bad user-experience. The fastest model, Token Dictionary, processes more than 70 queries within a second, making it the ideal choice in terms of user-experience.

\begin{table}
	\begin{center}
		\renewcommand{\arraystretch}{1.4}
		\setlength\tabcolsep{3pt}
		\begin{tabular}{lcccc}
			\hline\noalign{\smallskip}
			Method & Training & ~~Mean prediction~~ & Peak Memory~~ & Model 
			\vspace{-.1cm}
			\\
			 & time (s) & time (ms) & Usage (MB) & Size (MB)\\
			\noalign{\smallskip}
			\hline
			\noalign{\smallskip}
			Naive Bayes  & 21    & 17.7 & 1061 & 1.2\\
			Linear SVM   & 361  & 147.7 & 1250 & 1.9\\
			LSTM         & 12083 & 34.9 & 3232 & 294.1\\
			Random Forest& 41    & 71.9 & 4583 & 20.8\\
			Token Dict   & 16    & 12.8 & 143 & 0.7\\
			\hline
		\end{tabular}
		\vspace{.2cm}
	\caption{Runtime and memory consumption.}
	\label{tab:runtime}
	\end{center}
\end{table}

\subsection{Memory Consumption}
The last evaluation concerns memory. Although all models currently fit on the 8GB machine, it is important that the algorithms still work when more data are available in the future. As shown in Figure~\ref{tab:runtime}, token dictionary is memory-friendly and consumes less than 150 MB even for the largest training set we have. In comparison, random forest and LSTM model requires 4583MB and 3232MB. These results indicate that random forest and LSTM models might need additional memory on a workstation if more training data are available, while token dictionary is able to capture the mapping relationship between tokens and classes using a rather small amount of memory. Note that the models are compressed.



%
\subsection{Discussion}
\label{sec:discussion}

In the evaluation of the classifiers under scrutiny we haven seen that the token dictionary features very good classification results combined with favourable running time and memory consumption. A considerable part of the latter is probably also due to the fact that it has been implemented outside the scikit-learn framework. For example one notices a seven-fold difference in the peak memory usage of Naive Bayes compared to the token dictionary, which cannot be explained by the complexity of the method. Since the classification performance of the token dictionary exceeded the performance of the other methods we did not re-implement the other methods for a more accurate resource consumption comparison. Among further experiments we ran on this data set we evaluated the classification results when expanding abbreviations and observed that it does not bring a significant improvement. On the other hand, including 2-grams and 3-grams in the token dictionary does improve the classification as inter-token dependencies can be captured with little additional effort.
Thanks to the evaluation of the accuracy compared to the number of training files used, we have seen that our approach (regardless of the classification method) can produce good results already for relatively small data sets. More precisely, even if only 35 past projects are used for training the prediction results offer a high enough accuracy to reduce the workload of the engineers. For good performance we recommend to use around 85 projects for training (50\% of the data set available in our evaluation).

\section{Conclusion}
\label{sec:conclusion}
We modeled the substation signal name matching task as a classification problem and evaluated a set of common machine learning methods as well as a bespoke LST and token-based dictionary classifier on a data set built from past substation engineering project. 
Our proposed token dictionary method offers the fastest and most memory-efficient solution for the given task. Moreover, it gives the most accurate list of suggestions and competitive single-prediction results. 
A potential direction of future work concerns unseen tokens. Due to the nature of bag-of-word models, when encountering unseen tokens, it is impossible for these classifiers to convert these tokens into features, thus they will fail to capture the information in these tokens which can lead to low-quality predictions. One way to address this is to use some distance measure to find the closest known tokens to replace it.
%
%


\begin{thebibliography}{10}

\bibitem{aggarwal2012survey}
Aggarwal, C.C., Zhai, C.:
\newblock A survey of text classification algorithms.
\newblock Mining text data (2012)  163--222

\bibitem{bilenko2003adaptive}
Bilenko, M., Mooney, R., Cohen, W., Ravikumar, P., Fienberg, S.:
\newblock Adaptive name matching in information integration.
\newblock IEEE Intelligent Systems \textbf{18}(5) (2003)  16--23

\bibitem{rahm2001survey}
Rahm, E., Bernstein, P.A.:
\newblock A survey of approaches to automatic schema matching.
\newblock the VLDB Journal \textbf{10}(4) (2001)  334--350

\bibitem{hotho2005brief}
Hotho, A., N{\"u}rnberger, A., Paa{\ss}, G.:
\newblock A brief survey of text mining.
\newblock In: Ldv Forum. Volume~20. (2005)  19--62

\bibitem{mccallum1998comparison}
McCallum, A., Nigam, K.,  et~al.:
\newblock A comparison of event models for naive bayes text classification.
\newblock In: AAAI-98 workshop on learning for text categorization. Volume
  752., Madison, WI (1998)  41--48

\bibitem{pedregosa2011scikit}
Pedregosa, F., Varoquaux, G., Gramfort, A., Michel, V., Thirion, B., Grisel,
  O., Blondel, M., Prettenhofer, P., Weiss, R., Dubourg, V.,  et~al.:
\newblock Scikit-learn: Machine learning in python.
\newblock Journal of Machine Learning Research \textbf{12}(Oct) (2011)
  2825--2830

\bibitem{liaw2002classification}
Liaw, A., Wiener, M.,  et~al.:
\newblock Classification and regression by randomforest.
\newblock R news \textbf{2}(3) (2002)  18--22

\bibitem{caruana2006empirical}
Caruana, R., Niculescu-Mizil, A.:
\newblock An empirical comparison of supervised learning algorithms.
\newblock In: Proceedings of the 23rd international conference on Machine
  learning, ACM (2006)  161--168

\bibitem{joachims1998text}
Joachims, T.:
\newblock Text categorization with support vector machines: Learning with many
  relevant features.
\newblock Machine learning: ECML-98 (1998)  137--142

\bibitem{bottou2010large}
Bottou, L.:
\newblock Large-scale machine learning with stochastic gradient descent.
\newblock In: Proceedings of COMPSTAT'2010.
\newblock Springer (2010)  177--186

\bibitem{tang2015document}
Tang, D., Qin, B., Liu, T.:
\newblock Document modeling with gated recurrent neural network for sentiment
  classification.
\newblock In: EMNLP. (2015)  1422--1432

\bibitem{pennington2014glove}
Pennington, J., Socher, R., Manning, C.:
\newblock Glove: Global vectors for word representation.
\newblock In: Proceedings of the 2014 conference on empirical methods in
  natural language processing (EMNLP). (2014)  1532--1543

\bibitem{graves2005framewise}
Graves, A., Schmidhuber, J.:
\newblock Framewise phoneme classification with bidirectional lstm and other
  neural network architectures.
\newblock Neural Networks \textbf{18}(5) (2005)  602--610

\bibitem{bishop2006pattern}
Bishop, C.M.:
\newblock Pattern recognition and machine learning.
\newblock springer (2006)

\bibitem{kingma2014adam}
Kingma, D., Ba, J.:
\newblock Adam: A method for stochastic optimization.
\newblock arXiv preprint arXiv:1412.6980 (2014)

\bibitem{abadi2016tensorflow}
Abadi, M., Agarwal, A., Barham, P., Brevdo, E., Chen, Z., Citro, C., Corrado,
  G.S., Davis, A., Dean, J., Devin, M.,  et~al.:
\newblock Tensorflow: Large-scale machine learning on heterogeneous distributed
  systems.
\newblock arXiv preprint arXiv:1603.04467 (2016)

\end{thebibliography}
\end{document}